\documentclass[runningheads]{llncs}
\usepackage[T1]{fontenc}
\usepackage{graphicx,verbatim}
\usepackage{multirow}
\usepackage{booktabs}
\usepackage{subfigure}
\usepackage[dvipsnames]{xcolor}
\begin{document}
\title{CountPath: Automating Fragment Counting in Digital Pathology}
%
\author{Ana Beatriz Vieira\inst{1,2}\and
Maria Valente \inst{3,4}\and
Diana Montezuma\inst{5,6}\and Tomé Albuquerque\inst{4} \and Liliana Ribeiro\inst{5} \and Domingos Oliveira\inst{5} \and João Monteiro\inst{5} \and Sofia Gonçalves\inst{5} \and Isabel M. Pinto\inst{5} \and Jaime S. Cardoso\inst{3,4} \and Arlindo L. Oliveira\inst{1,2}}
\authorrunning{A. B. Vieira et al.}
%
\institute{Instituto Superior Técnico, Lisbon, Portugal \\
\email{anabeatrizvieira@tecnico.ulisboa.pt} \and
INESC-ID, Lisbon, Portugal \and
Faculdade de Engenharia da Universidade do Porto, Porto, Portugal \and 
INESC-TEC, Porto, Portugal \and
IMP Diagnostics \and
Cancer Biology and Epigenetics Group, Research Center of IPO Porto (CI-IPOP)/RISE@CI-IPOP (Health Research Network), Portuguese Oncology Institute of Porto (IPO Porto)/Porto Comprehensive Cancer Centre (Porto.CCC)}



\maketitle              
\begin{abstract}
Quality control of medical images is a critical component of digital pathology, ensuring that diagnostic images meet required standards. A pre-analytical task within this process is the verification of the number of specimen fragments, a process that ensures that the number of fragments on a slide matches the number documented in the macroscopic report. This step is important to ensure that the slides contain the appropriate diagnostic material from the grossing process, thereby guaranteeing the accuracy of subsequent microscopic examination and diagnosis. Traditionally, this assessment is performed manually, requiring significant time and effort while being subject to significant variability due to its subjective nature. To address these challenges, this study explores an automated approach to fragment counting using the YOLOv9 and Vision Transformer models. 
Our results demonstrate that the automated system achieves a level of performance comparable to expert assessments, offering a reliable and efficient alternative to manual counting. Additionally, we present findings on interobserver variability, showing that the automated approach achieves an accuracy of 86\%, which falls within the range of variation observed among experts (82--88\%), further supporting its potential for integration into routine pathology workflows.


\end{abstract}

\section{Introduction}

Digital Pathology has transformed medical diagnostics by automating image analysis, improving efficiency, and streamlining pathology workflows \cite{cancers16091686}. However, the adoption of these technologies has introduced new challenges in quality control, namely in ensuring that digital slides accurately reflect the biological material received. One key pre-analytical task is detecting and counting tissue fragments on pathology slides, crucial for preventing material loss and contamination. Traditionally, biomedical scientists perform this task manually by comparing histology slides with macroscopic reports, a process that, while effective, is time-consuming and prone to human error, especially with high slide volumes.
Accurate fragment detection and quantification are essential to verifying that the tissue analyzed corresponds precisely to what was received. Fragment variability in size, shape, and arrangement can complicate counting, and discrepancies between the observed and reported number of fragments trigger quality control reviews. These reviews help identify potential errors such as specimen misprocessing, tissue loss, or documentation issues, ensuring corrective actions are taken to maintain diagnostic accuracy.


Subjectivity is inherent in manual counting, as interpretations of fragment boundaries can vary between observers. For example, a fragment breaking during processing may prompt one observer to review the paraffin block, while another may consider the break insignificant. While such variability cannot be eliminated, automated approaches can help standardize assessments and reduce inconsistencies.
Currently, manual fragment counting is performed twice daily at IMP Diagnostics, taking 4–5 hours to process 1500–1600 slides. About 20–30 cases per day require review, with up to five needing corrective procedures like re-cutting. Although the percentage of cases needing review and correction is relatively low, this is relevant to maintaining a high-quality standard. This type of quality check is important across pathology laboratories, where accuracy and consistency directly impact patient outcomes.

\begin{figure}[!ht]
    \centering
    \includegraphics[width=0.9\textwidth]{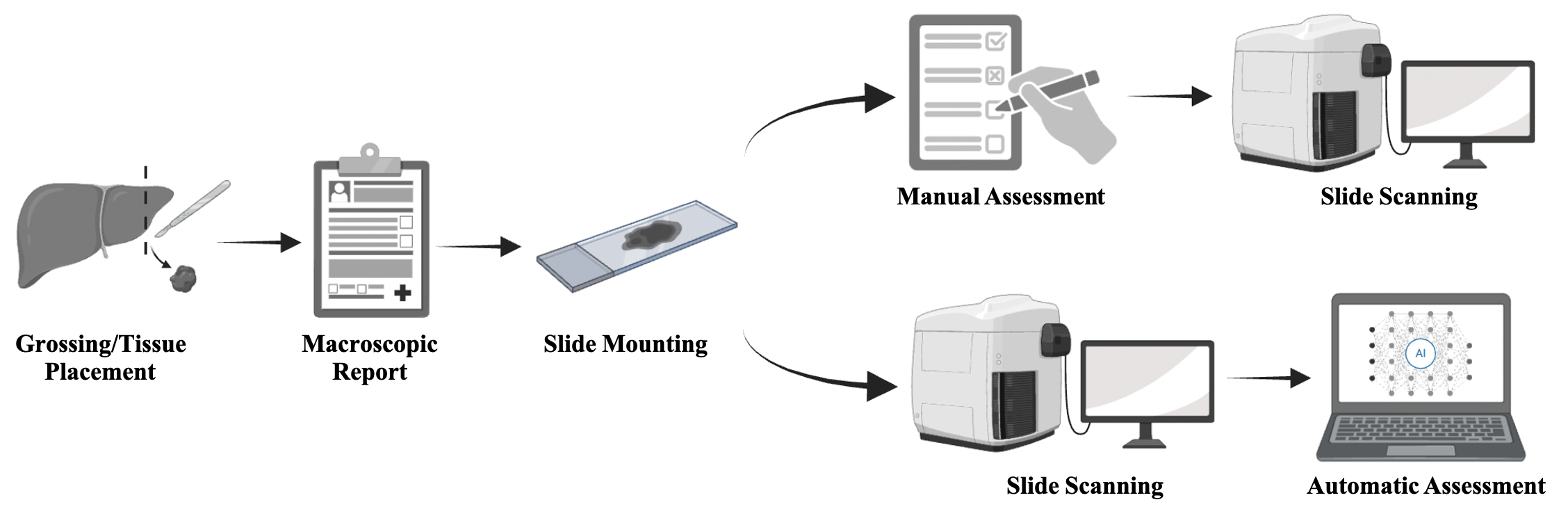}
    \caption{Illustration of manual and automatic assessment.}
    \label{fig:fig1}
\end{figure}

In prior work, Albuquerque et al. \cite{albuq} developed an automated system for detecting and counting fragments in colorectal biopsy whole-slide-images (WSIs) using both conventional machine learning and deep learning models, including YOLOv5 and Faster R-CNN. Their system improved accuracy and reduced manual workload by identifying discrepancies between slides and macroscopic reports.
Building on this foundation, we propose an advanced automated system for fragment detection and counting in digital pathology. Our approach aims to enhance accuracy, reduce workload, and improve diagnostic reliability. We validate its performance against manual counts from seven experts. 
Figure \ref{fig:fig1} illustrates the manual and automated assessment process.

\section{Methods}
This section describes the methodology, including dataset details, detection and counting methods for fragments and sets in WSIs, and a rejection option to exclude uncertain predictions. Finally, it describes the performed interobserver variability analysis used to assess consistency with human annotations, providing insights into reliability.

\subsubsection{Dataset:}

\begin{figure}[!ht]
    \centering
    \subfigure[1 set | 1 fragment]{%
        \includegraphics[width=0.31\textwidth]{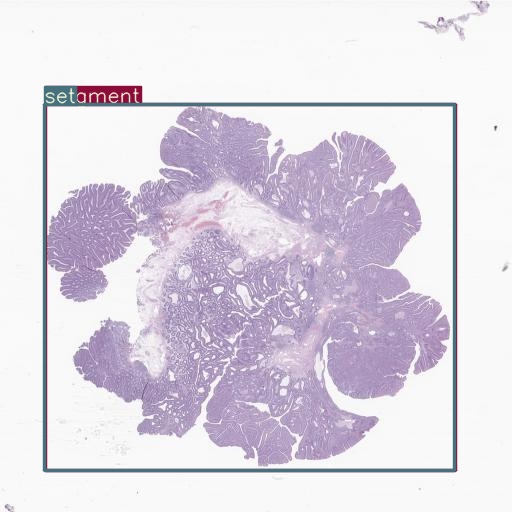}
        \label{fig:2a}}
    \subfigure[1 set | 3 fragments]{%
        \includegraphics[width=0.31\textwidth]{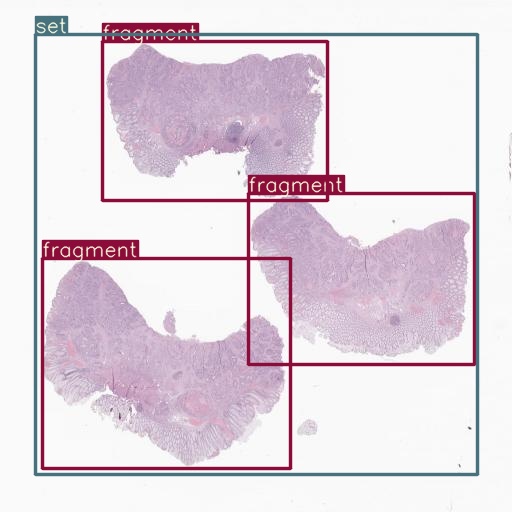}
        \label{fig:2b}}
    \subfigure[1 set | 3 fragments (x2)]{%
        \includegraphics[width=0.31\textwidth]{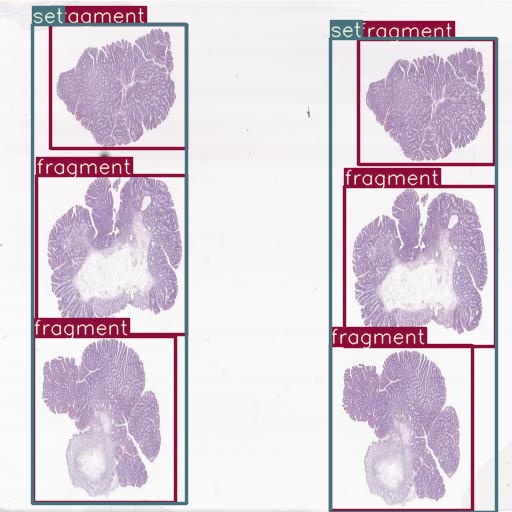}
        \label{fig:2c}}
    \caption{Representation of different tissue configurations in images. (a) One set containing a single fragment. (b) One set with three distinct fragments. (c) Two repeated sets, each containing three fragments of identical size and morphology.}
    \label{fig:fig2}
\end{figure}

The dataset comprises 3,253 WSIs from the IMP Diagnostics archive, digitized using three Leica GT450 WSI scanners at 40× magnification. This study specifically uses $1024 \times 1024$ thumbnails (low-resolution) images representations embedded within the pyramid-encoded WSI files--providing an efficient overview of the tissue layout \cite{FRAGGETTA201846}. Each histology slide could contain one set of different fragments (small pieces of tissue from biopsy or surgery procedures) or repeated sets of the same fragments, as shown in Figure \ref{fig:fig2}. These repeated sets are typically arranged to ensure that the tissue is adequately represented across multiple sections, allowing for better analysis.

\noindent Each case was evaluated by at least one pathologist and one biomedical scientist, and the number of fragments was compared with the original macroscopic report. Discrepancies were resolved through discussion between the evaluators or by consulting a third colleague. The dataset includes two types of annotations: spatial annotations, which specify bounding boxes, and numeric annotations, which indicate the number of fragments per set. Notably, all images with spatial annotations also have numeric annotations, ensuring that the entire dataset is numerically annotated. Cases with more than 9 fragments per set are labeled as class 10, reflecting the laboratory’s decision that an exact count is unnecessary for cases when the number of sets exceeds this value.

\subsubsection{Detection and Counting Models:}

We explored two complementary methods for counting fragments and sets in histopathology images: detection and classification. In the detection approach, fragment counts are derived from model predictions and refined through post-processing rules that prevent overlapping sets and ensure fragments are counted only within defined sets. The final count is obtained by dividing the number of fragments by the number of sets. If this ratio is not an integer, indicating a discrepancy, two strategies could be applied: (a) rounding the value to the nearest whole number or (b) flagging the sample for manual review. In contrast, the classification approach directly assigns images to predefined categories based on the number of fragments per set.

For detecting fragments and sets, we selected a state-of-the-art object detection model, YOLOv9, a convolutional neural network designed for object detection. It enhances earlier versions with an optimized architecture, improved feature extraction, and refined loss functions \cite{10.1007/978-3-031-72751-1_1}. Pre-trained on the MS COCO dataset, the model is fine-tuned on the target dataset, with input images annotated with bounding boxes and classes. For classification, we used the Vision Transformer (ViT), which processes images as patch-based tokens using self-attention mechanisms, similar to the one used in natural language processing NLP \cite{dosovitskiy2021an}. Pre-trained on ImageNet and fine-tuned on our dataset, ViT classifies images into predefined categories that corresponds to the fragment count.

\begin{figure}[!ht]
    \centering
    \includegraphics[width=0.98\textwidth]{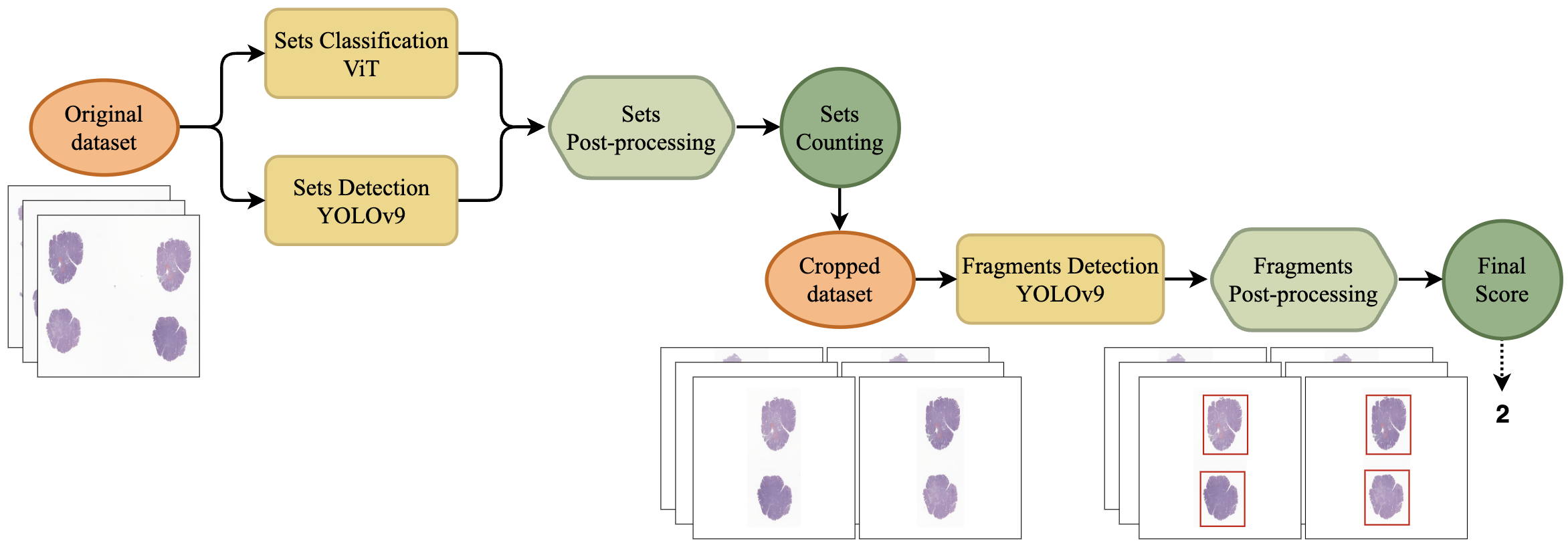}
    \caption{Overview of the proposed approach for detecting and counting fragments in WSIs. The process begins with the original dataset, where YOLOv9 is used for detecting sets. Then, a ViT-based model performs set classification/counting. Detected sets undergo post-processing and counting, generating a cropped dataset. Fragments are then detected using YOLOv9, post-processed, and the final score is determined.}
    \label{fig:fig3}
\end{figure}

We implemented a hybrid two-stage approach combining YOLOv9 and ViT to improve counting accuracy. In the first stage, both models are used to detect and count sets, with the ViT model imposing an upper bound on the number of sets detected by YOLOv9 to mitigate potential over-counting errors. Post-processing refines these detections by removing overlaps and filtering out low-confidence predictions, ensuring that only reliable detections are retained. Each detected set is then cropped to isolate a single set per region. In the second stage, these cropped images are processed by a (different) YOLOv9 model to detect fragments within each set. The final count is then refined through a post-processing step that ensures consistency across all cropped images derived from the same original image. Figure \ref{fig:fig3} illustrates the proposed approach.

\subsubsection{Rejection Option:}
In automated image analysis, particularly for complex medical imagery like histopathology slides, ensuring accurate classification is crucial for reliable downstream applications. While high-confidence predictions are ideal, machine learning models often produce varying confidence levels, which can lead to misinterpretations in precision-critical applications.
To address this challenge, introducing a "rejection option" has emerged as a valuable strategy for filtering out samples that do not have a reliable classification. This approach acknowledges that some samples may be inherently ambiguous, leading to misclassifications. By rejecting these cases, the model’s overall reliability improves. In our YOLOv9-based fragment and set detection, we implemented a simple integrity criterion: samples were rejected if the calculated fragment-to-set ratio was not an integer, indicating potential classification errors. In the hybrid approach, the rejection option is applied when the fragment counts in all cropped images from the same original image are inconsistent, ensuring reliability in the final count.

\subsubsection{Interobserver Variability:} \label{interobserver}

We selected a subset of 100 samples from the test set and asked a panel of seven biomedical scientists, with experience ranging from less than one year to over five years, to record the final score for each image. Given the inherent subjectivity in fragment counting, we assessed interobserver variability, which quantifies differences in classification among observers \cite{moons2023measuringagreementratersclassifying}.
To measure agreement, we computed the intraclass correlation coefficient (ICC), a standard metric for assessing reliability within a group of raters \cite{doi:10.2466/pr0.1966.19.1.3}. Additionally, we calculated Fleiss' Kappa, an extension of Cohen's Kappa, to evaluate inter-rater reliability or agreement for multiple raters \cite{fleiss,moons2023measuringagreementratersclassifying}. Higher values of these metrics indicate stronger agreement \cite{dataqualityevaluation2020}.

\section{Experiments}
In this section, we present the dataset composition, preprocessing and post-processing steps, along with a detailed description of the hyperparameters and training configurations used for the models. All experiments were conducted on a PowerEdge C41402 server with an NVIDIA 32GB Tesla V100S.

\subsubsection{Datasets and Preprocessing:}
The dataset was divided into train, validation, and test subsets, containing 2,053, 499, and 701 images, respectively. Each image was resized to either $512\times 512$ or $224\times 224$ pixels, depending on the model architecture. Following set detection, a new dataset was created by cropping these regions from the original images, resizing them to the model’s input size, and applying white padding. All images were normalized from their original pixel values (0–255) to a 0–1 range. Furthermore, to align with the pre-trained ViT network, the images were normalized using the specified mean and standard deviation for 224×224 images. Given the relatively limited dataset size, various data augmentation techniques were applied during training, further detailed below, to increase the robustness of the learning process.

\subsubsection{Post-processing:} The post-processing stage consists of two phases: set counting and fragment counting, each ensuring consistency and reliability in detection results. In set counting, the number of sets predicted by the ViT model serves as an upper bound, constraining YOLOv9 detections. Redundant bounding boxes are removed by discarding those with overlapping vertices, and remaining detections are sorted by confidence. The lower-confidence boxes are removed until the number of detections aligns with ViT’s output, ensuring consistency. Each set is then used to generate a cropped region. As a result, the number of cropped images corresponds to the sets identified in the original image, ensuring each set is isolated for further processing in the fragment detection stage.

In fragment counting, the process ensures consistency by verifying whether all cropped images derived from the same original image contain the same number of fragments. The count is considered valid only if all corresponding cropped regions yield the same result. In cases of discrepancy, the final score is determined based on the number of cropped images: if there are two images with differing counts, the lower value is selected; if there are more than two, the final score is chosen through majority voting.

\subsubsection{Implementation Details:}

We conducted all experiments using YOLOv9 and ViT-Base, implemented in PyTorch. YOLOv9 was trained for 200 epochs with a batch size of 32, using SGD with an initial learning rate of $10^{-2}$, which was reduced to $10^{-3}$, and a weight decay of $5 \times 10^{-4}$. Given the small dataset, data augmentation techniques--including vertical and horizontal flipping, translation, HSV adjustments, and mosaic augmentation--were applied. ViT-Base/32 was trained for 200 epochs with a batch size of 32, using AdamW with a learning rate of $10^{-5}$ and a weight decay of $5 \times 10^{-2}$. To enhance generalization, horizontal and vertical flipping were applied as data augmentation strategies.

\section{Results and Discussion}
This section presents the experimental results, offering a quantitative comparison of the ViT and YOLOv9 models alongside our proposed method for fragment and set counting. Additionally, we analyze interobserver variability in fragment counting and compare it to the best-performing automated model.

\subsubsection{Quantitative Analysis:}
We trained the YOLOv9-C model to detect fragments and sets, evaluating its performance on 701 samples with and without a rejection option. The rejection criterion was based on the ratio of detected fragments to sets: if this division did not yield an integer, the sample was rejected. The final score was computed as the ratio of detected fragments to sets, rounded to the nearest integer when the rejection option was not applied. We also implemented a ViT-based model for classifying fragments and sets and compared its results against YOLOv9. Additionally, we developed a method that combines YOLOv9 and ViT architectures to count sets and fragments. In this case, the rejection option is applied when the fragment counts in all cropped images from the same original image are inconsistent, ensuring reliability in the final count. A comparative analysis of these results obtained with these methods is presented in Table \ref{tab1}.

\begin{table}[h]
    \centering
    \caption{Performance evaluation of fragment and set counting using YOLOv9-C, ViT-B/32, and our proposed method. Results are presented both with and without rejection criteria, where the rejection percentage indicates the proportion of cases excluded. For the rejected cases, the evaluation metrics are computed without these samples. Precision, recall, and F1-score are weighted to account for class imbalances.}\label{tab1}
    \begin{tabular}{p{1.9cm} p{1.6cm} p{1cm} p{1cm} p{1.6cm} p{1.5cm} p{1.3cm} p{1.5cm}}
    \toprule
    \hfil \textbf{Model} & \hfil \textbf{Rejection} & \hfil \textbf{MAE} & \hfil \textbf{R$^2$} & \hfil \textbf{Accuracy} & \hfil \textbf{Precision} & \hfil \textbf{Recall} & \hfil \textbf{F1-Score} \\ 
    \midrule
    \hfil {YOLOv9-C} & \hfil No & \hfil 0.106 & \hfil 0.955 & \hfil 0.914 & \hfil 0.916 & \hfil 0.914 & \hfil 0.914 \\
    \midrule
    \hfil ViT-B/32 & \hfil No & \hfil 0.291 & \hfil 0.890 & \hfil 0.789 & \hfil 0.780 & \hfil 0.789 & \hfil 0.782 \\
    \midrule
    \hfil {Our method} & \hfil No & \hfil \textbf{0.084} & \hfil \textbf{0.965} & \hfil \textbf{0.932} & \hfil \textbf{0.932} & \hfil \textbf{0.932} & \hfil \textbf{0.931} \\
    \midrule \midrule
    \hfil YOLOv9-C & \hfil Yes(3.71\%) & \hfil 0.093 & \hfil 0.956 & \hfil 0.927 & \hfil 0.929 & \hfil 0.927 & \hfil 0.927 \\
    \midrule
    \hfil Our method & \hfil Yes(4.56\%) & \hfil \textbf{0.066} & \hfil \textbf{0.968} & \hfil \textbf{0.949} & \hfil \textbf{0.949} & \hfil \textbf{0.949} & \hfil \textbf{0.949} \\
    \bottomrule
    \end{tabular}
\end{table}

\noindent The YOLOv9-C model achieved an overall accuracy of 91.4\% when classifying all 701 samples without rejection. With 3.71\% of the samples rejected, the accuracy increased to 92.7\%, leaving 675 samples classified and 26 rejected. The decision to reject samples based on whether the fragments-to-sets ratio was an integer should be weighed in terms of its benefit to the final task. 

In comparison, the ViT-B/32 model achieved only 78.9\% accuracy. This lower performance is partly due to the ambiguity in defining fragments and the difference in training data—YOLOv9-C was trained with bounding box annotations, while ViT relied on image-level labels.
However, the ViT model exhibited outstanding performance in set classification, achieving 99.3\% precision and a 99.6\% F1-Score. This result underscores the effectiveness of ViT and attention mechanisms in accurately identifying sets, where the definition is clear and well-defined.

Our proposed method outperformed both models, achieving an accuracy of 93.2\% without rejection, which further improved to 94.9\% when 4.56\% of the samples were excluded. This result demonstrates that the rejection strategy effectively enhances reliability by filtering out ambiguous cases. These results underscore the advantages of integrating both architectures in a hybrid approach, leveraging the strengths of YOLOv9-C for precise fragment detection and ViT for robust set counting.

\subsubsection{Interobserver Variability Analysis:}
As mentioned, a group of seven observers was asked to annotate a subset of 100 images from the original test set. To assess the agreement between the observers, we calculated both the ICC and Fleiss' Kappa, two commonly used metrics for measuring inter-rater reliability. The ICC was 0.814, and Fleiss' Kappa was 0.74, indicating substantial agreement between the evaluators \cite{Landis1977TheMO}. Given this, we compared the performance of the automated method with the performance of the manual counting performed by these observers. It is important to note that the subset, although randomly selected, contains a high percentage of complex images that caused uncertainty or were difficult for the models to detect accurately in previous tests. The results from both the observers and the proposed method are presented in Table \ref{tab2}.

\begin{table}[h]
    \centering
    \caption{Comparison of the results obtained with our proposed method and through manual count by 7 observers, on a subset of 100 samples from the test set. Precision, recall, and F1-score are weighted to account for class imbalance.}\label{tab2}
    \begin{tabular}{p{1.8cm} p{1.6cm} p{1.55cm} p{1.65cm} p{1.65cm} p{1.65cm} p{1.6cm}}
    \toprule
    \hfil \textbf{Model} & \hfil \textbf{MAE} & \hfil \textbf{R$^2$} & \hfil \textbf{Accuracy} & \hfil \textbf{Precision} & \hfil \textbf{Recall} & \hfil \textbf{F1-Score} \\ 
    \midrule
    \hfil Observers & \hfil 0.210$\pm$0.02 & \hfil 0.90$\pm$0.02 & \hfil 0.849$\pm$0.02 & \hfil 0.779$\pm$0.08 & \hfil 0.723$\pm$0.06 & \hfil 0.735$\pm$0.07 \\
    \midrule
    \hfil Our method & \hfil 0.140 & \hfil 0.960 & \hfil 0.860 & \hfil 0.849 & \hfil 0.860 & \hfil 0.850 \\
    \bottomrule
    \end{tabular}
\end{table}

\noindent The results show that our method outperforms the manual counting by observers in multiple evaluation metrics. The automated approach achieves a lower Mean Absolute Error (MAE) and higher R², reflecting a strong correlation between predictions and ground truth values. Additionally, it achieves higher precision, recall, and F1-score, demonstrating improved reliability in detecting and counting fragments. With an accuracy of 86\%, within the observers' range (82–88\%), the automated approach proves to be a reliable alternative, potentially exceeding human performance in certain cases.

\section{Conclusion}
This work addressed the challenges of manual fragment counting in digital pathology by developing a hybrid method that integrates YOLOv9 and a Vision Transformer (ViT) model. Our approach was designed to enhance accuracy and reduce interobserver variability in this critical quality control task.

Compared to manual counting performed by seven experts, our proposed method achieved an accuracy of 86\%, surpassing the experts' average of 84.9\%, although by a margin that is not statistically significant for a 5\% significance level. These results demonstrate its potential to at least match and, in some measures exceed human performance in fragment detection and counting.

Our findings demonstrate that deep learning-based approaches, particularly the new method proposed, can improve pathology workflows by increasing accuracy, minimizing variability, and reducing the risk of diagnostic inconsistencies. By leveraging the strengths of both object detection and transformer-based models, our approach offers a robust and scalable solution for automated quality control in digital pathology.

    

\begin{credits}
\subsubsection{\ackname} 
This work was partially supported by national funds provided by Fundação para a Ciência e Tecnologia (FCT), under projects PRELUNA, PTDC/CCI-INF/4703/2021) and UIDB/50021/2020. In addition, IMP Diagnostics also provided funding support for this study.
We want to thank the biomedical scientists that also contributed to the interobserver variability study and/or data collection (Joana Guimarães, BSc; Mara Guedes, BSc; Cátia Gonçalves, BSc; Filipa Rebolo, BSc; Joana Ferreira, BSc; Vânia Leal, BSc).

\subsubsection{\discintname}
The authors have no competing interests to declare that are relevant to the content of this article.
\end{credits}

%
%
%
\bibliographystyle{splncs04}

%
\end{document}